\documentclass[runningheads]{llncs}
\usepackage[T1]{fontenc}

\newcommand{\node}{\textit{PhytoNode}}
\usepackage{float}
\usepackage{xcolor}
\usepackage{amsmath}
\usepackage{svg}
\usepackage{shellesc}
\usepackage{relsize}
\DeclareUnicodeCharacter{2212}{\ensuremath{-}}
\usepackage{textcomp}
\usepackage{multirow}
\usepackage{wrapfig}
\usepackage{subcaption}
\usepackage{graphicx}

\usepackage{multirow}
\definecolor{correction}{RGB}{255, 0, 0}

\begin{document}
%
\title{When Plants Respond: Electrophysiology and Machine Learning for Green Monitoring Systems}

%
%
\titlerunning{When Plants Respond}
%

\author{Eduard Buss\inst{1}\orcidID{0000-0001-6993-5873} \and
Till Aust\inst{1}\orcidID{0000-0003-2863-1341} \and  \\
Heiko Hamann\inst{1}\orcidID{0000-0002-2458-8289}}
\authorrunning{E. Buss et al.}
%

\institute{University of Konstanz, Department of Computer and Information Science, \\Konstanz, Germany \\
\email{\{eduard.buss, heiko.hamann\}@uni-konstanz.de}\\
}
%
\maketitle              
\begin{abstract} 
Living plants, while contributing to ecological balance and climate regulation, also function as natural sensors capable of transmitting information about their internal physiological states and surrounding conditions. 
This rich source of data provides potential for applications in environmental monitoring and precision agriculture. 
With integration into biohybrid systems, we establish novel channels of (physiological) signal flow between living plants and artificial devices. 
%
We equipped \textit{Hedera helix} with a plant-wearable device called \node, to continuously record the plant's electrophysiological activity. We deployed plants in an uncontrolled outdoor environment to map electrophysiological patterns to environmental conditions. Over five months, we have collected data that we analyzed using state-of-the-art and automated machine learning (AutoML).
Our classification models achieve high performance, reaching macro F1-scores of up to 95\% in binary tasks. AutoML approaches outperformed manual tuning and selecting subsets of statistical features further improved accuracy. 
%
Our biohybrid living system monitors the electrophysiology of plants in harsh, real-world conditions. 
This work advances scalable, self-sustaining, and plant-integrated living biohybrid systems for sustainable environmental monitoring. 




\end{abstract}
%
%
%

\section{Introduction}

Natural living plants are adaptive organisms equipped with smart sensing mechanisms that enable them to survive in diverse environmental conditions~\cite{volkov2012phytosensors}. Their adaptability makes them ideal candidates to integrate them in the development of bio-hybrid systems. 
%
They present various mechanisms that enable interactions with artificial components. For example, their directional growth in response to stimuli, called tropism, can help to grow living structures for human use~\cite{Hamann_2015}. Another key mechanism is plant communication that occurs internally between organs or externally with other plants and animals~\cite{Skrzypczak_2017}.
%
Understanding the principles underlying natural sensing, perception, and communication in plants could lead to novel living biohybrid solutions for environmental monitoring and interaction. We explore plant electrophysiology as a foundation for biohybrid sensing and phytosensing, using \textit{plant wearables} to measure and interpret electrical signals of the plant in response to environmental stimuli. 

Internal signaling occurs via hydraulic, chemical, or electrical pathways~\cite{Skrzypczak_2017}. Electrical signals are generated by ion movements (Ca\textsuperscript{2+}, K\textsuperscript{+}, Cl\textsuperscript{–}, H\textsuperscript{+}), which are regulated by channels, pumps and transporters to enable rapid responses~\cite{Johns_2021}. These membrane potential shifts initiate processes like photosynthesis or movement. For example, in \textit{Mimosa pudica}, mechanical stimulation triggers an action potential propagating at 44~mm/s, leading to leaf folding within 1~s as response~\cite{Hagihara2020,Johns_2021}.
Plants generate and propagate four distinct types of electrical potentials: action potential, variation potential, local electrical potential, and system potential. 
These signals differ in their triggers, duration, propagation range, intensity, and consistency. They travel through the phloem and xylem that are also responsible for transporting water and nutrients~\cite{Johns_2021,Li_2021}.
%
%
Analyzing these can be complex as they can overlap, forming composite signals that convey complex information about the plant's physiological state and environment. However, this superposition complicates interpretation. 
We use supervised machine learning (ML) to analyze these unknown signals. This leads to a \textit{living omni-sensor} (detecting diverse environmental signals via one biological system), where plants naturally respond to stimuli, and intelligent algorithms translate their signals into human-readable data. 
We have selected ivy (\textit{Hedera helix}) as our species due to its widespread distribution in Europe, adaptability to various habitats, evergreen nature, and climbing-associated morphological variability, making it a strong candidate for environmental monitoring applications~\cite{Metcalfe2005,buss_2024}.

Our work, previously supported by the EU-funded research project \textit{WatchPlant} until 2024, focuses on monitoring environmental conditions through plant physiology. During \textit{WatchPlant}, we developed an interactive, intelligent network of biohybrid sensor nodes capable of monitoring urban air pollution using a sustainable and environmentally friendly approach. 

We have introduced a novel design for a bio-hybrid sensor node, the \node, and demonstrated its functionality in controlled and semi-controlled experimental setups~\cite{buss_2024}. As key contribution of this paper, we moved our experiments out of the lab to outdoor environments, hence, making the experiments uncontrolled. These are harsh conditions for both the hardware and our signal analysis approach. The hardware is exposed to weather and operates off the grid. Our analysis of the electrophysiological signal is challenged by the plant being exposed to multiple co-existing uncontrolled stimuli of the outdoor environment. We validate the measurement performance, weatherproof casing of the \node, and its energy harvesting capabilities. 

We continue with an overview of related research on phytosensing (Sec.~\ref{Sec. Related Work}). Subsequently, we describe the outdoor experimental setup (Sec.~\ref{sec. experimental_setup}) and the applied machine learning approach (Sec.~\ref{sec. machine_learning}). The results (Sec.~\ref{Sec. Results and Discussion}) present electrical potential measurements and classifier performances. Finally, we discuss key findings, limitations, and future directions (Sec.~\ref{sec. conclusion}).

\section{Related Work} \label{Sec. Related Work}
Monitoring plant physiology is usually done to monitor the plant's health or to gain insights for plant biology. Our focus is on phytosensing, that is, monitoring plant physiology to learn about the environmental conditions through the plant. 

Plant wearable devices enable monitoring of plant physiological parameters to allow timely interventions, such as irrigation adjustments. An example of such a device was developed by 
Fiorello et al. \cite{fiorello2021plant}, who introduced a plant-inspired miniature device that features microscopic hooks to enable stable attachment to leaf surfaces. Integrated with sensors for parameters, such as temperature and humidity, the system allows for minimally invasive monitoring of the leaf microclimate. Additionally, they demonstrated a proof-of-concept prototype, \textit{MiniBot}, which can traverse the leaf surface by engaging its hooks through a soft fluidic multiphase actuator activated by near-infrared laser stimulation.
 
The electrical potential of plants functions as a complex yet universal carrier of information, providing valuable insights into plant health and responses to environmental conditions. This has led to the application of machine learning approaches to identify stimulus-specific patterns. For example, Bhadra et al.~\cite{bhadra2023multiclass} exposed tomato and cabbage plants to three environmental chemical stimuli: ozone, sulfuric acid, and sodium chloride. They classified the electrical responses of the plants using AdaBoost applied to 15 statistical features, achieving an F1\mbox{-}score of 0.71. Various performance metrics and classifiers were evaluated on the collected imbalanced dataset.

Similarly, Dolfi et al.~\cite{Dolfi2015} investigated the electrophysiological response of \textit{Ligustrum texanum} and \textit{Buxus macrophilla} to ozone exposure. They applied ozone concentrations of up to 240 \textmu g/m\textsuperscript{3}, a typical alarm threshold set by regulatory agencies. Based on signal derivatives in combination with correlation waveform analysis, they achieved detection accuracies of up to 92\%.

A common challenge in ML is the choice of appropriate preprocessing steps as well as the choice of the classifier and its hyperparameters. Its choice often depends on experience, trial and error, or rule of thumbs. In automated machine learning (AutoML), we are able to automatically compose and parameterize ML algorithms to optimize a given metric~\cite{Yao2018}. Current state-of-the-art frameworks are \textit{auto-sklearn}~\cite{Feurer2015,Feurer2020},  GAMA~\cite{gijsbers2019gama} and Naive AutoML~\cite{Mohr2022}. 
Aust~et~al.~\cite{Aust2024} propose a generic framework for classifying ozone and wind stimuli based on \textit{ts\_fresh}~\cite{Christ2018} and Naive AutoML library.
They chose Naive AutoML because it yields similar performance by significantly less computational time.  


\section{Methods}
We measured electrical potential variations in \textit{Hedera helix} under uncontrolled outdoor conditions to identify environmentally induced responses. Four \linebreak \node\mbox{-}equipped plants were placed near a monitoring station (Sec.~\ref{sec. experimental_setup}). Continuous measurements were recorded, statistical features extracted, and measurements labeled based on environmental conditions. Various machine learning models were applied for high-accuracy classification (Sec.~\ref{sec. machine_learning}).


\subsection{Experimental Setup} \label{sec. experimental_setup}
\begin{wrapfigure}{r}{0.4\textwidth}
    \vspace{-2.0cm}
	\centering
	\includegraphics[width=0.39\textwidth]{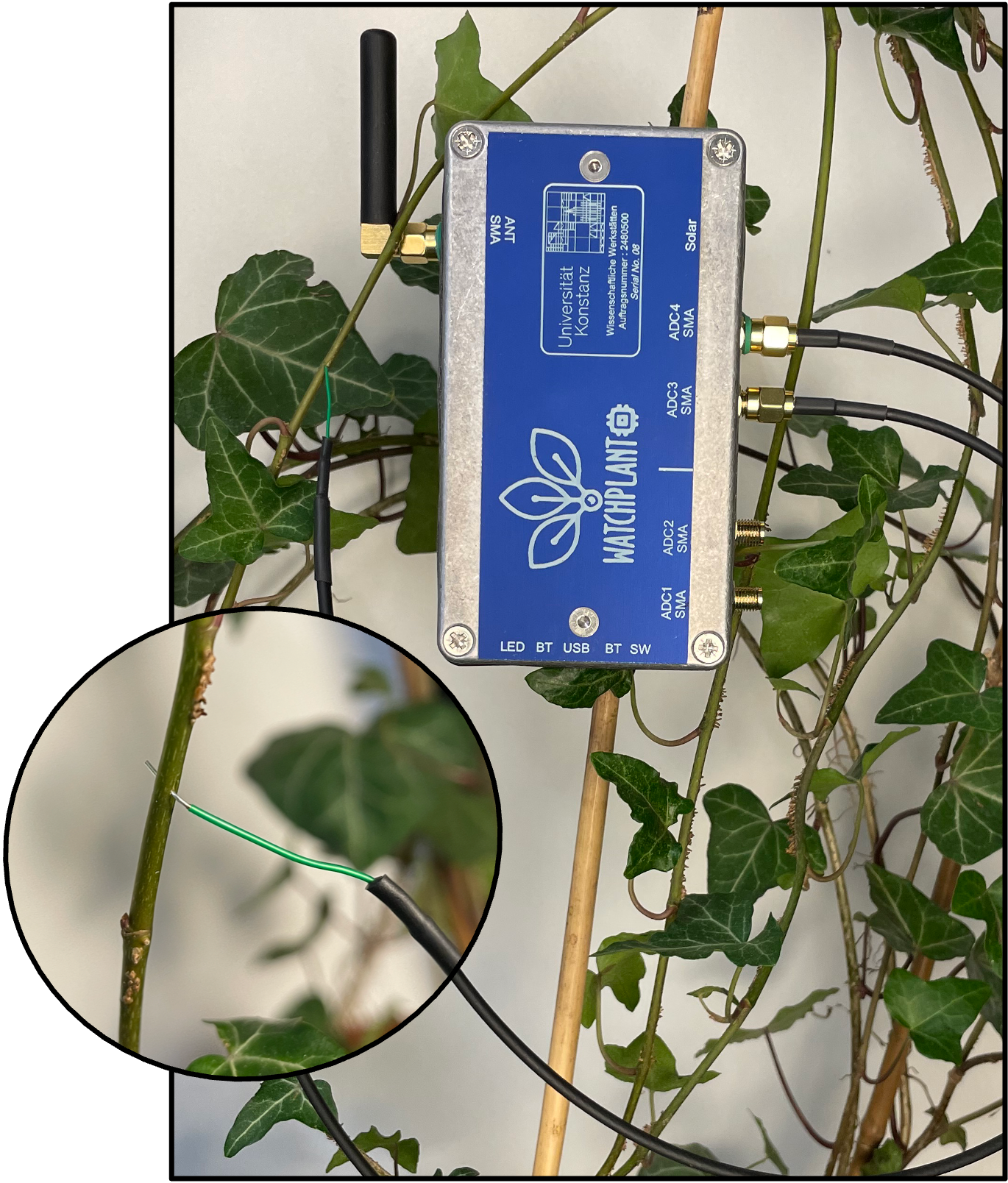}
	\caption{PhytoNode deployed.} 
    \label{fig phytonode}
    \vspace{-0.8cm}
\end{wrapfigure}



We monitor the electrical potential of ivy using our custom-developed \textit{\node} (Fig.~\ref{fig phytonode}). It is a self-sustaining sensor node powered by a LiPo battery, which is recharged using a solar panel. It transmits the recorded data and ML outputs via Bluetooth Low Energy (BLE) to a data sink or directly to nearby pedestrians through a dedicated mobile app~\cite{buss_2022}. The \textit{\node} functions as a plant-wearable device, measuring the electrical potential of the plant using a pair of silver-coated electrodes~\cite{buss_2024}. One electrode is inserted at the lower end of the stem, just above the soil, while the other is placed either in the same stem or in the petiole of a leaf on the same stem, maintaining a distance of 30~cm to 60~cm, depending on the plant's size. We assume that measuring at the stem provides a more holistic view of the plant's physiological state, while measurements at the leaf offer a localized view of one leaf. We protect the electronics from electromagnetic interference with coaxially sheathed electrodes and an aluminum housing. The electrical potential is sampled at a frequency of about 200~Hz and transmitted to a Raspberry~Pi~4B, which uploads the data for further processing.

\begin{figure}[b]
	\centering
	\includegraphics[width=0.8\textwidth]{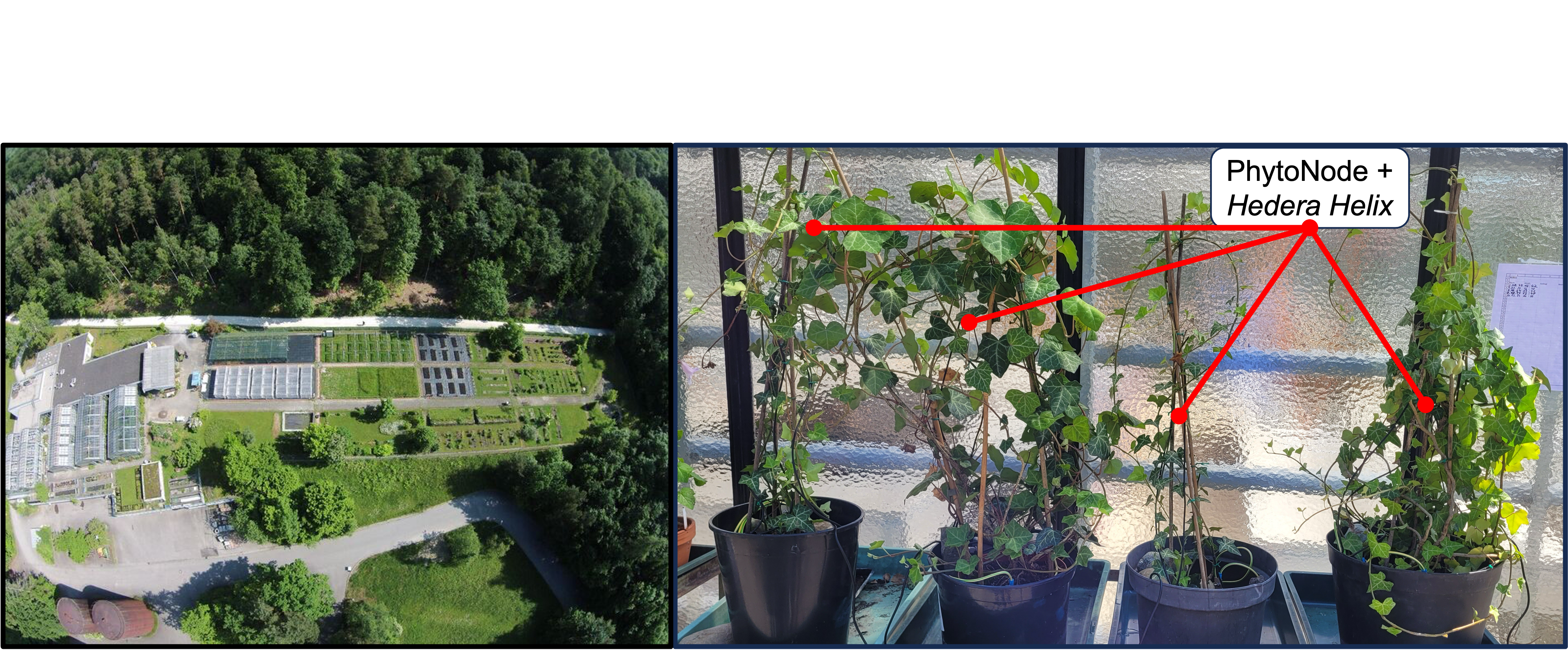} 
	\caption{Experimental setup with four ivies (right) in an outside location at the University of Konstanz (Germany) botanical garden (left, drone image top view).}
    \label{fig botanical garden}
\end{figure}

We collect electrical potential measurements from four \textit{Hedera helix} of varying sizes and ages (Fig. \ref{fig botanical garden}) to identify species-related patterns that extend individual plant characteristics. The plants are positioned outdoors, close to a greenhouse in the botanical garden of the University of Konstanz. 
Data acquisition spans from July~5,~2024, to November~18,~2024. We analyze these measurements in relation to environmental conditions monitored by the garden's measurement station provided by Adolf Thies GmbH\&Co.~KG, located approximately 20~m away. 
Environmental parameters include wind speed~[m/s], wind direction~[°], air temperature~[°C], relative humidity~[\%], solar irradiance~[W/m\textsuperscript{2}], precipitation~[mm], and dew point temperature~[°C], all recorded at a sampling frequency of 0.1~Hz.  
\subsection{Machine Learning Analysis} \label{sec. machine_learning}
Our objective is to map the electrical potential measurements to typical environmental conditions, such as warm/cold and day/night.
The general pipeline for analyzing these measurements is as follows: (1)~Preprocessing the raw measurements. 
(2)~Time window selection under specific environmental conditions. 
(3)~Statistical feature calculation of each time window. 
(4)~Training of multiple classifiers using either all or selected features. 
(5)~We compare the results of manually crafted ML (ManualML) approaches with processing pipelines identified by automated machine learning (AutoML). 

\noindent\textit{(1) Preprocessing Raw Measurements.} 
The plant response was sampled at a frequency of approximately 200~Hz. First, we downsampled the time series to 1~Hz for resource efficiency by applying a mean filter over 1~s intervals. This smoothed the data and reduces the impact of missing samples due to hardware communication issues. Days with less than 80\% data coverage were excluded to maintain data reliability. If mentioned in the results, we interpolate the remaining days with missing samples for a continuous, uniformly sampled time series. We use time-based interpolation which establishes linear relationships between data points based on the timestamp~\cite{pandas}. After preprocessing, the resulting dataset consists of 216 days of electrical potential measurements. 
Again, if stated, the time series is z-score normalized using $z = \frac{x - \mu}{\sigma}$ where $z$ is the normalized sample, $x$ is the raw sample, $\mu$ the mean of the time series, and $\sigma$ is the standard deviation of the series. This transformation normalizes electrical potentials for comparability across plants and experiments, reducing physiological and experimental variations. Environmental measurements and downsampled electrical potential measurements are available online~\cite{buss_2025_dataset}.

\noindent\textit{(2) Time Window Selection Under Specific Environmental Conditions.} 
We use the environmental measurements described in Sec.~\ref{sec. experimental_setup} to divide electrophysiological measurements into categories, referred to as classes. For example, the distinction between day and night is based on solar irradiance measurements, using a defined threshold (Table~\ref{tab env_thresholds}). Time intervals with solar irradiance values above 50~W/m$^2$ are classified as day, while those below are classified as night. As temperature and wind speed follow  typical day-night cycles, we use data exclusively from 8~am to 8~pm for these two conditions. The 
thresholds depend on 
specific environmental conditions, the monitoring objectives, and the type of plant used. Our key design decision here was to create a setup that allows us to showcase the approach. Generally, the thresholds influence the difficulty of the classification task and would usually depend on requirements of the application. 

\begin{table}[t]
	\centering
	\caption{Eight extracted classes based on environmental features and thresholds.} 
    \label{tab env_thresholds}
	\begin{tabular}{lcc} 
		\hline
        Classes & Environmental Features & Threshold\\
        \hline
        Day - Night & Solar irradiance & 50 W/m2\\
        Rain - Dry & Precipitation & 0 mm \\
        Warm - Cold &  Air temperature & 25 °C \\
        Windy - Calm &  Wind speed & 1.25 m/s \\
        \hline
	\end{tabular}
\end{table}

\noindent\textit{(3) Feature Calculation.}
The time series data that meet the class criteria in \textit{(2)} are extracted from the down-sampled data described in \textit{(1)} and labeled with the respective class. These time series are then segmented into 1-h intervals to calculate features. The choice of a 1-hour window balances sensitivity to both rapid and gradual fluctuations in the measured differential potential while maintaining timely inference regarding environmental conditions. Nonetheless, additional analysis will be undertaken to identify the shortest feasible interval that still allows for reliable environmental assessments. Using the Python library \textit{tsfresh}, we compute over 700 statistical features for each interval, which are used for training and testing. 
We use an 80\%/20\% split between training and testing datasets, with 20\% of the training data reserved for validation.
To ensure numerical stability and comparability across features, we apply min-max normalization to the features that is given by $x_{\text{i, norm}} = \frac{x_i - \min(\textbf{x})}{\max(\textbf{x}) - \min(\textbf{x})}$ where $x_i$ represents the current sample, and $\textbf{x}$ the entire set of one feature. The gathered datasets are highly imbalanced (e.g., few time intervals with rain). To avoid the worst case that the classifier ignores the minor class, we up-sample the minor classes in the training set using Synthetic Minority Over-sampling Technique (SMOTE). SMOTE generates new samples between an original sample and neighboring samples (\textit{k}=5) combined with a random factor~\cite{chawla_2002}. However, synthetic data might not represent real-world scenarios or can lead to ambiguous samples.

\noindent\textit{(4) Training the classifiers. }
We classify the data using five distinct classifiers, including linear, non-linear, and deep learning approaches, which are briefly described below. A random forest classifier (RF) is an ensemble learning algorithm that builds multiple decision trees (here 256 trees) during training and merges their outputs to improve accuracy and reduce overfitting. 
Naive Bayes (NB) is a classification algorithm based on Bayes' theorem, which calculates the probability of a class given the input features.
It is simple, efficient, and particularly effective for classification tasks where the data shows an underlying Gaussian distribution. 
K-Nearest Neighbors (KNN) is a non-parametric algorithm that classifies data points based on the majority class of their $k$ (here $k$ = 5) closest neighbors in the feature space. 
Linear Support Vector Machine (SVM) finds the optimal hyperplane to separate data points of different classes. It maximize the margin between the hyperplane and the nearest data points (support vectors) of each class, making it effective for linearly separable data. 
A Multi-Layer Perceptron (MLP) is a type of artificial neural network composed of an input layer, one or more hidden layers, and an output layer, where each layer is fully connected to the next. We use an architecture with an input layer depending on the number of used features and three hidden layers of 50, 50, and 25 neurons, respectively. The output layer depends on the number of classes to classify. Further, the ReLU activation function, the \textit{Adam} optimizer, a batch size of 32 and an adaptive learning rate is used. 
The classifiers are trained using both the complete feature set and a selected subset to reduce dimensionality and eliminate redundant or potentially noisy information. The selection is conducted based on their mutual information (MI) that quantifies feature–class dependency. As previously discussed, we expect the dataset to be highly imbalanced due to the unequal distribution of environmental conditions. Consequently, classification accuracy may provide a misleading assessment of model performance. Instead, we use the macro F1-score as a more robust evaluation metric, as it considers the minority and majority class equally. 
All classifiers are trained on ten independent stratified shuffle splits of the training data. We use the Python library \textit{scikitlearn} for implementing the classifiers and feature selection algorithm \cite{scikitlearn_2011}.



\noindent\textit{(5) Automated Machine Learning.}
The selected classifiers include a diverse set of linear, nonlinear, and deep learning models. The chosen hyperparameters are based on prior experience and commonly used values in the literature. However, given the large search space of hyperparameter combinations, alternative classifiers, or preprocessing techniques that were not explored may potentially increase the performance. To address this, we compare the above approach with AutoML pipelines, which systematically explore a broader solution space to identify optimal classification pipelines.
The Naive AutoML library~\cite{Mohr2022} searches in \textit{sci-kit learn}~\cite{scikitlearn_2011} classifier and preprocessing implementations for a ML pipeline that optimizes a given metric (e.g., maximizing macro F1-score). 
The selection algorithm works greedy in two phases:
(a)~the best algorithm combination of preprocessors and classifiers is found by iterating through all possible combinations (discarding invalid combinations) using the algorithm's default parameter; 
(b)~the hyperparameters are optimized through random search for a pre-defined number of optimization steps. 
The best found pipeline is trained on the full dataset and returned. 
This library is easily configurable by setting timeout limits, maximum hyperparameter optimization iterations, or choosing the to be optimized metric. 
In our study, we choose 1024~hyperparameter optimization iterations with an early stopping for 100~rounds of no improvement. 
We do not set a timeout interval and select the macro F1-Score as metric to be maximized.


\section{Results and Discussion} \label{Sec. Results and Discussion}
\begin{figure}[t]
	\centering
    \includegraphics[width=1\linewidth,]{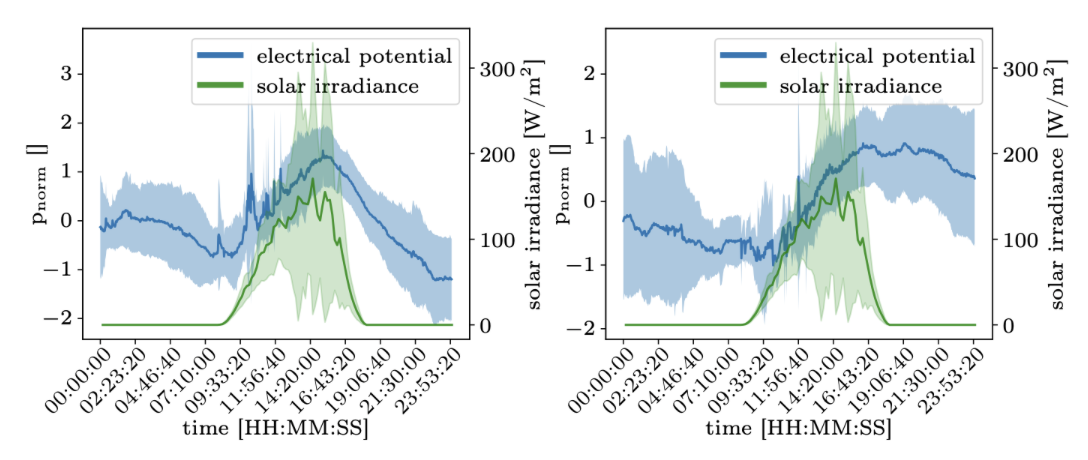}
	\caption{Normalized electrical potential measurements of six consecutive days. The daily mean (solid line) and standard deviation (shaded  area) are presented for measurements at the leaf (left) and measurements at the stem (right) in blue. Additionally, the solar irradiance during the same period is shown in green.}
    \label{fig normalized_week}
\end{figure}

We start by showing daily z-score normalized electrical potential measurements collected over the period from November~5, 2024, to November~12, 2025, including six consecutive days of data. On these days, sunrise was around 8~am and sunset around 6~pm (see green line in Fig.~\ref{fig normalized_week}).  For any missing values caused by hardware failures or external events, interpolation was used to ensure daily time series of equal length. The normalized data was then used to compute the daily mean (solid lines) and standard deviation (shaded areas), visualizing the daily dynamics of the plant's electrical response at the stem and the leaf. Additional visualizations of differential potential dynamics are available in \cite{buss_2024,Dolfi2015}. 

Our analysis follows a progressive approach, starting with qualitative and relative observations before moving on to a more sophisticated method. The electrical potential remains relatively stable and smooth overnight but rises to higher and more variable levels during the day. Consequently, the electrical potentials exhibit a clear day–night pattern.

Next, we split the data set into four binary classification tasks: day vs. night, calm vs. wind, rain vs. dry and warm vs. cold. We measured a total of 43,760 samples (here labeled sequences of 1~h) with both channels and all four Ivy plants combined. 
Since labeling distinguishes only between tow classes (e.g., \textit{day} and \textit{night} based on a \textit{solar irradiance} threshold), the same sample may belong to multiple classes. For example, the same sample may be classified as both \textit{cold} and \textit{dry} if both conditions are simultaneously met.
The dataset imbalance is caused by uneven occurrence of 
weather conditions. During the measurement period, colder temperatures were dominant, making the \textit{cold} class 79.4\% of the cold-warm (6,164 vs. 1,600 samples) 
dataset. Longer nights resulted in the \textit{night} class accounted for 61.0\% of the day-night (5,306 vs. 8,292 samples) dataset. Precipitation was rare, with the dry class at 95.9\% of the rain-dry (636 vs. 14,916 samples) 
dataset. Similarly, calm conditions dominated, with the \textit{calm} class at 93.6\% of the calm-wind (6,408 vs. 438 samples) 
dataset. To address this imbalance, techniques like up-sampling and appropriate classification metrics are necessary to prevent overfitting and improve 
accuracy.

The classification performance is evaluated using the macro F1-score, as shown in Table~\ref{tab: stem_f1}. This table also provides a comparative performance analysis between manually configured classifiers and the AutoML framework, which optimizes classification pipelines. 
Overall, the manually configured RF classifier and the AutoML-selected pipeline achieved the highest F1-scores across all classification tasks, with only minor deviations of up to 4.66\% (Leaf, Cold~-~Warm). When considering the average macro F1-score across all classification tasks, RF provides slightly superior performance at 90.7\%, compared to 89.6\% obtained by AutoML (Table~\ref{tab: stem_f1}, Stem). Following RF and AutoML, the MLP, SVM, KNN, and NB classifiers performed in descending order, with average macro F1-scores of 83.4\%, 78.9\%, 57.9\%, and 54.7\%, respectively. The performance differences become more distinct when analyzing class-wise accuracies. For instance, in distinguishing between wind and calm conditions using stem data, NB correctly classified 72.2\% of the minority class (wind) but only 59.8\% of the majority class (calm). In contrast, the AutoML (NOR + QDA) pipeline has a more balanced performance by correctly classifying also 72.0\% of the wind instances while achieving 100\% accuracy in classifying calm.
\begin{table} [t]
	\centering
	\caption{Macro F1-score average for stem and leaf measurements of both ManualML and AutoML classifer. The best classification results are highlighted in bold (NOR: Normalizer, QDA: Quadratic discriminant analysis, ETC: Extra tree classifier, RFC: Random forest classifier, PCA: Principle component analysis, VT: Variance threshold). }
	\label{tab: stem_f1} 
	\begin{tabular}{|c|c|l|c|c|c|c|c|c|c|c|c|} 
        \cline{3-7} 
         \multicolumn{2}{c|}{} &Class & Wind - Calm & Day - Night & Rain - Dry  & Cold - Warm \\
        \hline
        \multirow{7}{*}{\rotatebox{90}{\parbox{2 cm}{\centering Stem}}}&
        \multirow{4}{*}{\rotatebox{90}{\parbox{2 cm}{\centering ManualML}}} 
        &NB [\%]  & 31.90 $\pm$ 1.43 & 63.41 $\pm$ 1.41 & 60.02 $\pm$ 0.15 & 63.40 $\pm$ 1.20 \\
        &&KNN [\%] & 49.14 $\pm$ 0.53 & 69.43 $\pm$ 0.66 & 53.65 $\pm$ 0.19 & 59.30 $\pm$ 0.64 \\
        &&MLP [\%] & 80.51 $\pm$ 2.40 & 82.08 $\pm$ 2.34 & 88.50 $\pm$ 1.34 & 82.35 $\pm$ 1.51\\
        &&SVM [\%] & 73.75 $\pm$ 1.17 & 79.58 $\pm$ 0.54 & 84.87 $\pm$ 0.96 & 77.27 $\pm$ 0.41 \\
        &&RF [\%]  & 87.91 $\pm$ 0.85 & 92.80 $\pm$ 0.36 & \textbf{93.78 $\pm$ 0.56} & \textbf{88.48 $\pm$ 0.65}\\ \cline{2-7}
         & \multirow{2}{*}{\rotatebox{90}{\parbox{0.8 cm}{\centering Auto\\ML}}}&Pipeline & NOR + QDA & ETC & NOR + QDA & VT + PCA + ETC \\
        &&Score [\%]& \textbf{90.83 $\pm$ 0.00} &\textbf{93.79 $\pm$ 0.22} & 89.26 $\pm$ 0.00 & 84.69 $\pm$ 0.28 \\
        \hline
        \hline
        \multirow{7}{*}{\rotatebox{90}{\parbox{2 cm}{\centering Leaf}}}&
        \multirow{4}{*}{\rotatebox{90}{\parbox{2 cm}{\centering ManualML}}} 
        & NB [\%]  & 61.36 $\pm$ 1.50 & 55.84 $\pm$ 0.67 & 64.44 $\pm$ 0.29 & 60.42 $\pm$ 0.51 \\
        &&KNN [\%] & 52.85 $\pm$ 0.56 & 66.56 $\pm$ 0.40 & 56.31 $\pm$ 0.31 & 62.05 $\pm$ 0.95 \\
        &&MLP [\%] & 74.71 $\pm$ 5.96 & 79.66 $\pm$ 2.34 & 88.62 $\pm$ 1.08 & 79.14 $\pm$ 2.02\\
        &&SVM [\%] & 67.21 $\pm$ 1.24 & 75.54 $\pm$ 0.60 & 82.43 $\pm$ 0.64 & 75.45 $\pm$ 0.60 \\
        &&RF [\%]  & 87.31 $\pm$ 0.69 & 91.21 $\pm$ 0.68 & \textbf{93.65 $\pm$ 0.49} &\textbf{ 89.64 $\pm$ 0.69}\\ \cline{2-7}
        &\multirow{2}{*}{\rotatebox{90}{\parbox{0.8 cm}{\centering Auto\\ML}}}&Pipeline & QDA & RFC & PCA + QDA & NOR + PCA + QDA \\
        &&Score  [\%] & \textbf{90.83 $\pm$ 0.00} &\textbf{92.65 $\pm$ 0.19} & 89.26 $\pm$ 0.00 & 84.98 $\pm$ 0.22 \\
        \hline
	\end{tabular}
\end{table}
The classification F1-Score show only small differences between the measurements of the leaf and the stem. This is in contrast to \cite{Aust2024} where the leaf channel yielded much better results. This may be because the observed stimuli occur over longer timescales, changing more slowly, allowing their effects to fully propagate throughout the entire plant.

\begin{figure}[t] 
    \centering
    \begin{subfigure}[t]{6 cm}
        \centering
        \includegraphics[width=5.6 cm]{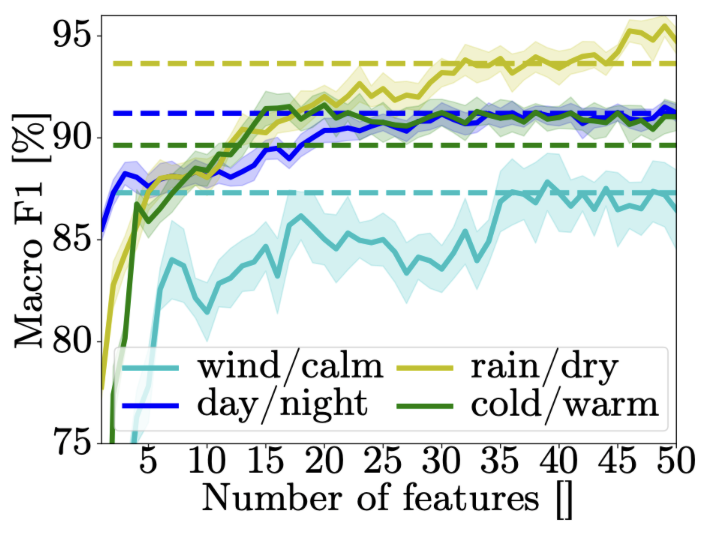}
        \caption{}
        \label{fig:feature selection results}
    \end{subfigure}
    \hfill
    \begin{subfigure}[t]{6 cm}
        \centering
        \includegraphics[width=6 cm]{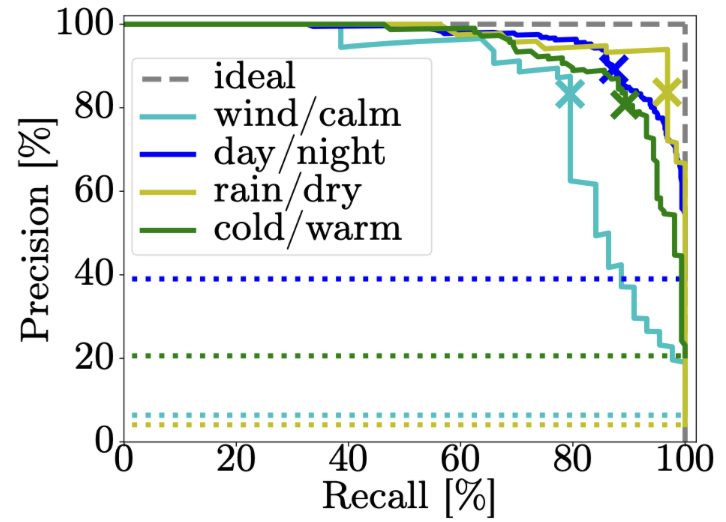}
        \caption{}
        \label{fig: PR_Curve}
    \end{subfigure}
    \caption{Leaf classifier performance. (a) RF classifier's average macro F1-score (solid line), standard deviation (shaded area), and maximum F1-score with all features (dashed line). (b) Precision-Recall curve (solid line) for all binary tasks using the best performing feature subset, with baseline (dotted), ideal classifier (gray, dashed) and best classifier performance (crosses).}
    \label{fig:main}
\end{figure}

Next, we analyze whether all features are necessary for high F1-scores using the RF classifier (highest average performance). The feature selection algorithm from Sec.~\ref{sec. machine_learning} is applied with an 80/20 train-test split and ten stratified shuffle splits on the training data. Fig.~\ref{fig:feature selection results} presents the mean and standard deviation of macro F1-scores for the top 50 features across all binary classification tasks. We exceeded the highest F1-scores obtained using all features with fewer features across all tasks. For wind vs. calm, the previous maximum (87.3\%) was surpassed with 36 features, peaking at 87.8\% ± 1.4\% with 39 features. In day vs. night, 36 features exceeded the prior best (91.21\%), reaching 91.5\% ± 0.3\% with 49 features. For rain vs. dry, 32 features outperformed the previous 93.7\%, achieving 95.5\% ± 0.5\% with 49 features. Cold vs. warm required the fewest, surpassing 89.6\% with 13 features and reaching 91.6\% ± 0.7\% with 20 features. The used features are available online~\cite{buss_2025_dataset}.
The precision-recall curve evaluates classifier performance based on the threshold probability for the minority class, making it ideal for imbalanced datasets. See Fig.~\ref{fig: PR_Curve} for each binary task, including the baseline (dotted) and ideal classifier (gray, dashed). The baseline, set by the minority class proportion (e.g., 20.6\% in cold-warm), represents random classifier performance. The area under the curve (AUC) quantifies classifier performance with rain-dry (97\%) leading, followed by day-night (96\%), cold-warm (94\%), and wind-calm (85\%). All best performing classifier using the subset are located at the top right corner of the PR-Curve (crosses, Fig.~\ref{fig: PR_Curve}), thus showing a good tradeoff between precision and recall. For example, the classifier of wind-calm achieved a recall of 0.79 and a recall of 0.83. In other words, 83\% of the minority class predictions were correct while capturing 79\% of the actual minority class.

\section{Conclusion} \label{sec. conclusion}
We monitored \textit{Hedera helix} for five months in outdoor conditions using our living biohybrid approach and recorded variations in their electrophysiology and environmental factors. Our goal is to develop a biohybrid living sensor for scalable, sustainable environmental monitoring. To analyze plant electrophysiological responses, we developed various linear, non-linear, and deep learning models and compared them to AutoML-generated pipelines. A~key challenge were the highly imbalanced datasets due to irregular environmental conditions. While we could have designed threshold-based workarounds, the challenge of imbalanced data is inherent and expected in the domain of living plant sensing. To improve classifier reliability given the high cost of real-world experiments, appropriate performance metrics and data augmentation techniques are essential. Manually configured classifiers varied in performance, while AutoML consistently achieved high F1-scores. As expected, AutoML is preferable as it explores a broader search space and yields high-performing classifiers. 
Initially, all classifiers were trained on over 700 features, which can introduce irrelevant or noisy variables, thus potentially hindering performance. Therefore, we analyzed classification performance with an increasing number of selectively chosen features and found that better performance can be achieved with fewer, more carefully selected features.
%

The classifier's generalizability may be limited by several factors: (1)~The study used only four \textit{Hedera helix} plants, introducing bias as physiological patterns may vary due to individual variability and across species. (2)~Data was collected from June to November 2024 in Konstanz, Germany 
using ivies, meaning patterns may differ across species, seasons, and geographic locations. (3)~Environmental feature thresholds were arbitrarily defined for labeling. 
To address these limitations, we aim to extend data collection throughout the entire year and include a larger number of plants to create a more diverse and generalizable dataset. Currently, data analysis is performed offline. As a next step, we plan to implement the classifiers directly on the \node, enabling real-time classification and transmission of results via BLE to nearby pedestrians or other recipients for immediate environmental awareness. Its self-sustaining design, weatherproof construction, and wireless communication capabilities enable to monitor the electrophysiology across groups of plants distributed in urban or agricultural environments (e.g.,  vineyards). This supports scalable real-time plant-health monitoring and early intervention by authorities or farmers. 

Overall, we provided a systematic evaluation of the electrophysiology of \textit{Hedera helix} within the context of environmental monitoring in harsh outdoor conditions.
We have developed a complete bio-hybrid pipeline for living machines, from experimental data acquisition, via preprocessing, to data analysis. 

\bibliographystyle{splncs04}
\bibliography{bibliography}

\end{document}